\documentclass[letterpaper]{article} 
\usepackage[preprint]{aaai2027} 
\usepackage[hyphens]{url}  
\usepackage{graphicx} 
\usepackage{natbib}  
\usepackage{caption} 
\usepackage{algorithm}
\usepackage{mathtools}
\usepackage{algorithmic}
\usepackage{subfiles}
\usepackage{amsmath}
\usepackage{amssymb}
\usepackage{xcolor}
\usepackage{booktabs}
\usepackage{newfloat}
\usepackage{listings}
\DeclareCaptionStyle{ruled}{labelfont=normalfont,labelsep=colon,strut=off} 
\floatstyle{ruled}
\newfloat{listing}{tb}{lst}{}
\floatname{listing}{Listing}

\usepackage{booktabs}

\title{Socialized Detector Learning: Trajectory-Guided and Reciprocal Distillation for Heterogeneous Object Detectors}
\author{
    Weihao Li\textsuperscript{\rm 1},
    Yunqi Zhu\textsuperscript{\rm 8},
    Zhihe Fan\textsuperscript{\rm 4}, 
    Ruipu Zhao\textsuperscript{\rm 2}, 
    Boan Tao\textsuperscript{\rm 3}, 
    Xinjie Yao\textsuperscript{\rm 5}, 
    Yan Fan\textsuperscript{\rm 6},\\
    Pengfei Zhu \textsuperscript{\rm 2,7}
}
\affiliations{
\textsuperscript{\rm 1}School of New Media and Communication, Tianjin University, Tianjin, China\\
\textsuperscript{\rm 2}School of Artificial Intelligence, Tianjin University, Tianjin, China\\
\textsuperscript{\rm 3}School of Computer Science and Technology, Tianjin University, Tianjin, China\\
\textsuperscript{\rm 4}School of Sports Training, Tianjin University of Sport, Tianjin, China\\
\textsuperscript{\rm 5}Faculty of Information Engineering and Automation,
Kunming University of Science and Technology, Kunming, China\\
\textsuperscript{\rm 6}School of Electronic Science,
National University of Defense Technology, Hunan, China\\
\textsuperscript{\rm 7}School of Automation, 
Southeast University, Nanjing, China\\
\textsuperscript{\rm 8}School of Computer Science and Engineering,
University of New South Wales, Sydney, Australia

}

\begin{document}

\maketitle

\begin{abstract}
Object detection knowledge is fragmented across independently trained, heterogeneous detectors with complementary category supports. In socialized learning, this knowledge resides in a society, and learning aims to evolve the society collectively through exchange. However, aggregation-based socialization does not explicitly plan transfer order, whereas progressive multi-teacher distillation considers order but remains a one-way student enhancement in a shared category space. Building on Socialized Learning, we formulate Socialized Detector Learning (SDL) for heterogeneous, category-specialized object detectors and propose Trajectory-Guided and Reciprocal Distillation (TGRD).TGRD estimates directed operational Inter-Detector Transfer Difficulty (IDTD) from held-out feature-alignment residuals, precomputes a fixed score table, and greedily constructs a carrier trajectory. Along the trajectory, knowledge is progressively consolidated into a union-category carrier and then returned to experts through reciprocal transfer. A conditional proxy-certificate analysis shows that, under stated assumptions, the progressive certificate is no larger than an aggregated-target counterpart. On MS COCO with four heterogeneous experts and two carrier initializations, final carriers outperform epoch-matched simultaneous aggregation controls by 2.6 AP in both settings. Reciprocal detectors attain 20.8--28.4 AP on previously unsupported categories while remaining within 1.3 AP of original expert-specific performance. These results support order-aware progressive consolidation followed by reciprocal transfer as a viable mechanism for detector-society evolution.
\end{abstract}


\section{Introduction}


\begin{figure}[t]
    \centering
    \includegraphics[width=\columnwidth]{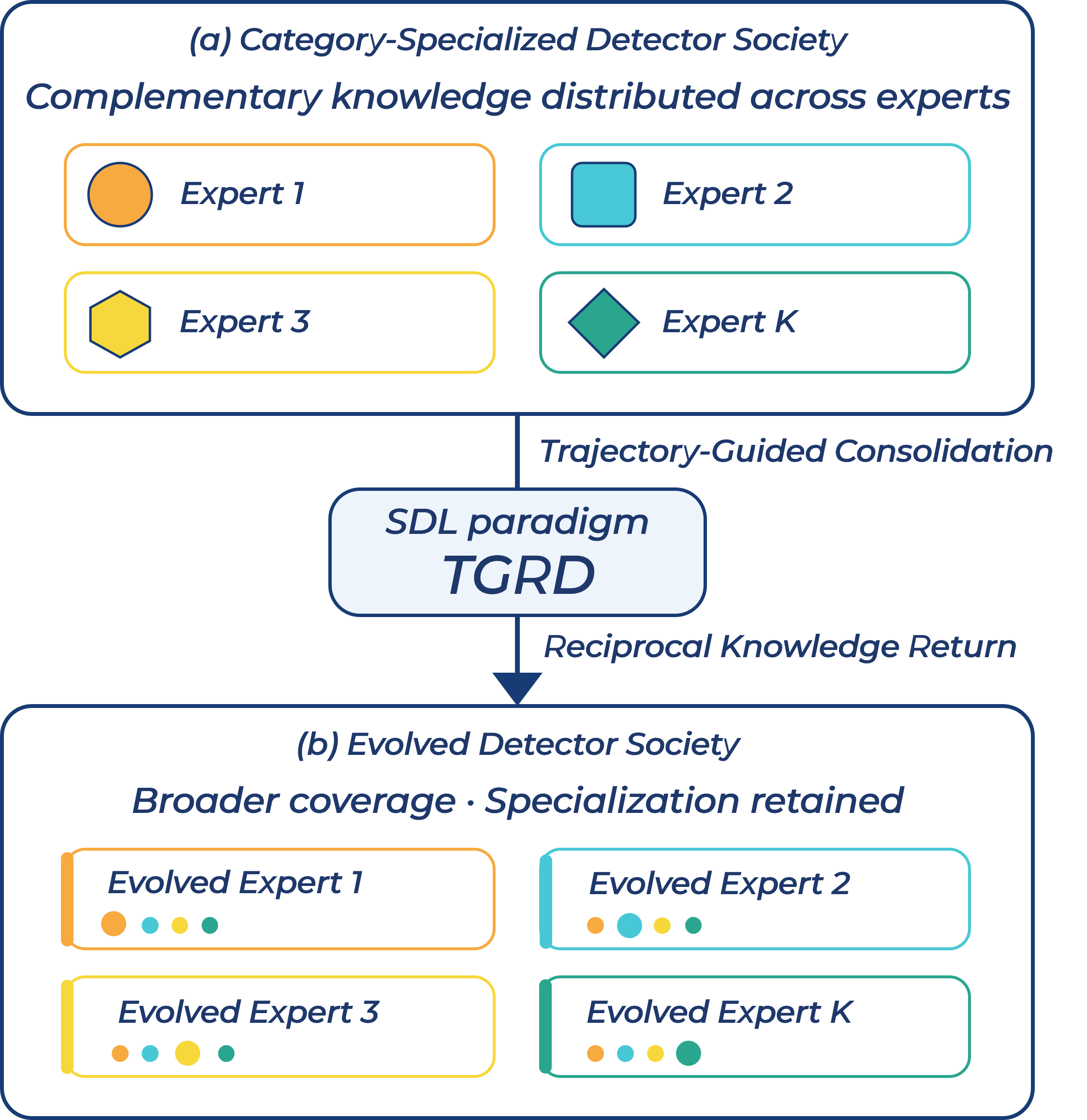}
    \caption{Detector-society evolution under SDL. TGRD progressively consolidates complementary expertise and reciprocally transfers it back, broadening category coverage while preserving specialization.}
    \label{fig:sdl_overview}
\end{figure}

Knowledge in object detection systems is often distributed across independently trained detectors with different category coverage, architectures, and data. 
Figure~\ref{fig:sdl_overview} illustrates the detector-socialization setting: distributed expertise is consolidated and redistributed to broaden category coverage while retaining detector-specific specialization.
Continual object detection~\cite{shmelkov2017incremental} updates one model along a temporally given task sequence, while conventional multi-teacher distillation transfers multiple teachers into a target student. Neither directly addresses the evolution of a heterogeneous detector society.

Two prior works directly motivate our formulation but operate at different boundaries. Socialized Learning (SL)~\cite{yao2024socialized} provides the society-level perspective that agents can improve through knowledge exchange. Its MASC framework realizes category-partitioned classification socialization through aggregation-based distillation and reciprocal altruism. However, MASC organizes distributed knowledge through aggregation and does not explicitly model or plan the order of knowledge transfer among agents. MTPD~\cite{cao2023learning} demonstrates that multiple detector teachers can be progressively distilled in sequence to improve a lightweight detector. It considers transfer order, but remains a one-way student-enhancement scheme rather than a framework for society evolution.

These limitations raise two questions:
\begin{enumerate}
    \item \emph{How can directed transfer difficulty guide a progressive trajectory among heterogeneous detectors?}
    
    \item \emph{How can reciprocal transfer broaden category coverage while preserving detector specialization?}
\end{enumerate}

To address these questions, 
we specialize the socialized-learning perspective to heterogeneous, category-specialized object detectors and refer to this setting as Socialized Detector Learning (SDL).
Within SDL, we develop \emph{Trajectory-Guided and Reciprocal Distillation} (TGRD). For the first question, TGRD derives directed operational Inter-Detector Transfer Difficulty (IDTD) scores from held-out feature-alignment residuals, precomputes a fixed score table, and greedily constructs a carrier trajectory. For the second, the carrier progressively consolidates distributed expertise along this trajectory while expanding to the union category vocabulary; its consolidated knowledge is then returned to individual experts through reciprocal transfer. The updated detectors constitute the evolved society.

For the progressive phase, we provide a conditional proxy-certificate comparison with an abstract aggregated-target alternative: under the stated conditions, the progressive certificate is no larger than its aggregated counterpart. Experiments on MS COCO with four heterogeneous experts and two carrier initializations support the complete TGRD configuration. The final carriers outperform epoch-matched simultaneous aggregation controls by 2.6 AP in both settings, while the reciprocal detectors attain 20.8--28.4 AP on previously unsupported categories and remain within 1.3 AP of their original expert-specific performance.

Our contributions are summarized as follows:
\begin{itemize}
\item 
We formulate socialization among heterogeneous detectors as acquiring complementary category capabilities while retaining specialization.

\item 
We propose TGRD, combining IDTD-guided planning, progressive union-category consolidation, and reciprocal transfer, with a conditional proxy-certificate analysis.

\item 
COCO experiments show higher final-carrier AP than Avg-FPN KD and broader coverage after reciprocal updates, with limited specialization change.

\end{itemize}


\section{Related Work}

\textbf{Continual Object Detection.}

Continual object detection updates a detector over sequential tasks while preserving previously learned categories. A common formulation uses a temporal update chain, where the detector trained on earlier tasks teaches the detector updated on new tasks.
Beginning with incremental detection without storing old data~\cite{shmelkov2017incremental}, later methods reduce forgetting
through selective and inter-related distillation~\cite{peng2021sid},
importance-aware classification and localization
transfer~\cite{feng2022overcoming}, intra- and inter-class
distillation~\cite{kang2023alleviating}, and cross-stage knowledge
alignment~\cite{mo2024bridge}.

However, their underlying structure remains centered on a single evolving detector. Knowledge is transferred mainly from the previous model to the current model, and the learning order is dictated by task arrival rather than by compatibility among models. This differs fundamentally from Socialized Detector Learning, where knowledge is distributed across independently trained expert detectors that may vary in categories, architectures, data, or detection paradigms. SDL therefore requires compatibility-aware consolidation among heterogeneous experts, rather than merely preserving old knowledge along a single temporal update chain. 

\textbf{Knowledge Distillation for Object Detection.}

Early works extend KD to object detection through output distillation and feature imitation~\cite{chen2017learning,li2017mimicking,wang2019distilling}. Later methods refine what and where to distill by exploiting instance-level, decoupled, focal, global, and localization-aware knowledge~\cite{dai2021general,guo2021distilling,yang2022focal,zheng2022localization}. Recent studies further explore masked or scale-aware feature reconstruction, heterogeneous teacher-student alignment, DETR-specific distillation, cross-head prediction mimicking, and automated distillation policy search~\cite{huang2023masked,zhu2023scalekd,lao2023unikd,chang2023detrdistill,wang2024crosskd,li2024detkds}.

Most detection KD fixes teacher--student roles and transfer direction, with
research centered on distillation signals and alignment. MTPD~\cite{cao2023learning}
is the closest related setting: it uses a non-symmetric adaptation cost to order multiple teachers, but assumes a shared category space and one-way transfer to a lightweight student. TGRD likewise uses directed ordering, but operates across independently trained detectors with different category supports, progressively expands a carrier toward their union category space, and returns the consolidated knowledge to the original detectors. Its endpoint is therefore an updated detector society rather than a single enhanced student.

\textbf{Federated and Decentralized Model Aggregation.}

Federated and decentralized learning combines models trained by distributed
clients under privacy, communication, non-IID data, system heterogeneity, and
topology constraints. Representative mechanisms include matched parameter
averaging~\cite{wang2020federated}, ensemble
distillation~\cite{lin2020ensemble}, Bayesian
ensembling~\cite{chen2020fedbe}, prototype and layer-wise posterior
aggregation~\cite{tan2022fedproto,liu2024fedlpa}, hierarchical aggregation for
federated detection~\cite{jia2024adaptive}, and gossip-based decentralized
aggregation~\cite{hu2022spread}. They follow a distributed optimization
protocol, combining client updates through server or peer-to-peer
communication.

SDL differs in its objective and interaction semantics. Like federated learning, SDL may involve detectors trained independently on distributed data. However, SDL treats these detectors as a society of specialized experts whose knowledge should be exchanged, consolidated, and redistributed so that both the society and its members can evolve. Under this view, the central issue is not only how to aggregate distributed updates, but how to organize knowledge transfer among heterogeneous experts according to their compatibility, specialization, and reciprocal benefit. The present TGRD instantiation assumes centralized access to detector features and associated training data; privacy-preserving optimization and communication efficiency are outside the scope of this work.


\section{Method}

\begin{figure}[t]
    \centering
    \includegraphics[width=\columnwidth]{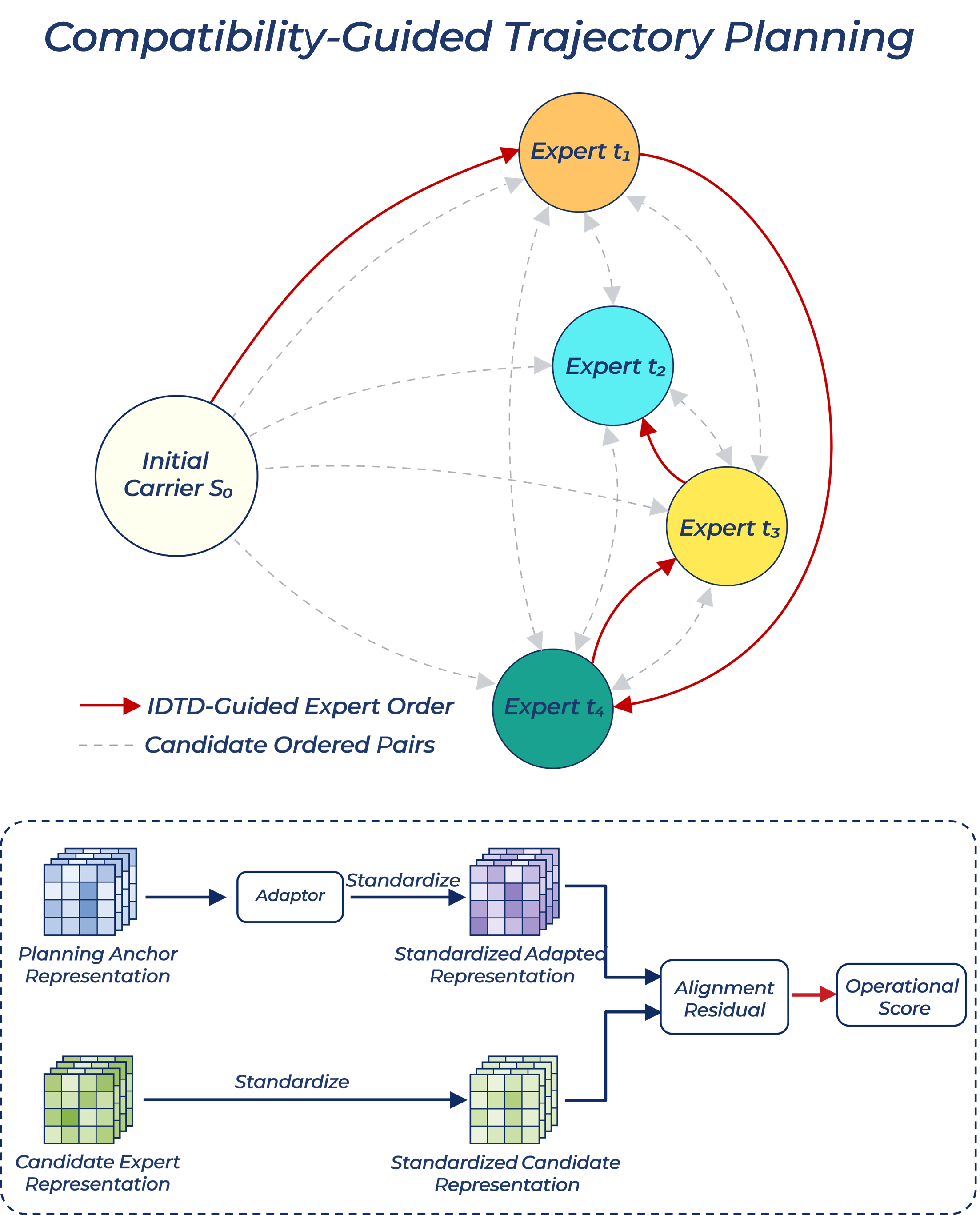}
    \caption{Compatibility-guided carrier trajectory planning. Held-out alignment residuals define a fixed table of directed operational scores $\widehat{D}(A,B)$ over ordered anchor--candidate pairs, from which greedy selection determines the expert order and carrier trajectory.}
    \label{fig:trajectory_planning}
\end{figure}
\label{sec:method}

\subsection{Socialized Detector Learning}

In real-world deployments, object detection knowledge is often distributed
across a society of independently trained expert detectors. These experts may
be heterogeneous in category coverage, architecture, training data, or
detection paradigm. We formulate \emph{Socialized Detector Learning} (SDL) as a
learning paradigm in which a detector society evolves collectively through
knowledge exchange. Let
\begin{equation}
    \mathcal{T}^{(r)}
    =
    \left\{
        t_1^{(r)},\ldots,t_K^{(r)}
    \right\}
    \label{eq:society}
\end{equation}
denote the society at evolution round \(r\). Each member induces an
expert-specific transfer target \(g_i^{(r)}\), abstracting transferable
knowledge arising from its category coverage, architecture, data, and
detection paradigm. One round of socialized learning is written as
\begin{equation}
    \mathcal{T}^{(r+1)}
    =
    \Phi_{\mathrm{SDL}}
    \!\left(
        \mathcal{T}^{(r)};\Omega^{(r)}
    \right),
    \label{eq:sdl_update}
\end{equation}
where \(\Omega^{(r)}\) specifies the knowledge-exchange protocol and
\(\Phi_{\mathrm{SDL}}\) updates the society from the exchanged knowledge. The
goal is an evolved society whose members acquire complementary capabilities
while retaining their specialized expertise.

\subsection{TGRD: Compatibility-Guided Carrier Trajectory Planning}

At round \(r\), TGRD receives \(\mathcal{T}^{(r)}\) and initializes a
standalone knowledge carrier \(S_0^{(r)}\). For conciseness, we write
\(\mathcal{T}=\{t_1,\ldots,t_K\}\) and \(S_0\) in this subsection. Let
\(\mathcal{C}_0\) be the categories initially supported by \(S_0\), and let
\(\mathcal{C}_i\) be those supported by \(t_i\). We use
\begin{equation}
    \mathcal{C}_{\cap}
    =
    \mathcal{C}_0
    \cap
    \bigcap_{i=1}^{K}\mathcal{C}_i
    \ne \varnothing,
    \qquad
    \mathcal{C}_{\cup}
    =
    \mathcal{C}_0
    \cup
    \bigcup_{i=1}^{K}\mathcal{C}_i
    \label{eq:category_spaces}
\end{equation}
as the common probe categories and the final society-wide category vocabulary,
respectively. 
All pairwise operational planning scores are computed using the same
two nonempty, finite, disjoint probe-image sets
\(\mathcal D_{\mathrm{fit}}^\cap\) and
\(\mathcal D_{\mathrm{eval}}^\cap\), constructed over
\(\mathcal C_\cap\).
This common probe is used only for compatibility
estimation and does not restrict the categories transferred during carrier
training. Figure~\ref{fig:trajectory_planning} summarizes the directed
compatibility estimation and the resulting carrier trajectory.

\paragraph{Carrier trajectory and progressive label space.}
TGRD constructs a permutation \(\pi\) of \(\{1,\ldots,K\}\), which defines
the carrier path
\begin{equation}
    \mathcal{P}_{\pi}:
    \quad
    S_0
    \xrightarrow{\,t_{\pi(1)}\,}
    S_1
    \xrightarrow{\,t_{\pi(2)}\,}
    \cdots
    \xrightarrow{\,t_{\pi(K)}\,}
    S_K .
    \label{eq:carrier_path}
\end{equation}
The arrow labels identify the expert supplying supervision at each stage. Let
\(\mathbf{c}_0=(c_{0,1},\ldots,c_{0,|\mathcal{C}_0|})\) and
\(\mathbf{c}_i=(c_{i,1},\ldots,c_{i,|\mathcal{C}_i|})\) be fixed ordered
category-name lists. Category names are unique within each list, and shared
categories use the same canonical name across detectors. Starting from
\(\boldsymbol{\kappa}_0=\mathbf{c}_0\) and
\(\mathcal{K}_0=\mathcal{C}_0\), stage \(k\) performs
\begin{equation}
    \begin{aligned}
        \Delta\boldsymbol{\kappa}_k
        &=
        \left[
            c\in\mathbf{c}_{\pi(k)}
            \,\middle|\,
            c\notin\mathcal{K}_{k-1}
        \right],
        \\
        \boldsymbol{\kappa}_k
        &=
        \boldsymbol{\kappa}_{k-1}
        \mathbin{\Vert}
        \Delta\boldsymbol{\kappa}_k,
        \qquad
        \mathcal{K}_k
        =
        \operatorname{set}(\boldsymbol{\kappa}_k).
    \end{aligned}
    \label{eq:progressive_categories}
\end{equation}
Here \(\Delta\boldsymbol{\kappa}_k\) preserves the order inherited
from \(\mathbf c_{\pi(k)}\), \(\parallel\) denotes list concatenation,
and \(\operatorname{set}(\cdot)\) returns the underlying set of
category names.
Thus \(\boldsymbol{\kappa}_{k-1}\) is a prefix of
\(\boldsymbol{\kappa}_k\): the corresponding class-specific output blocks are
copied from \(S_{k-1}\), and only the appended suffix blocks are newly
initialized. Exact category-name matching aligns expert and carrier outputs,
and \(\mathcal{K}_K=\mathcal{C}_{\cup}\).

\paragraph{Inter-Detector Transfer Difficulty.}
For an ordered pair \((A,B)\), \(A\) is the current planning anchor and \(B\)
is a candidate expert. The notation \(A\to B\) describes the intended
evolution of the carrier representation toward \(B\), while \(B\) supplies
knowledge during the ensuing carrier update. We define the latent
\emph{Inter-Detector Transfer Difficulty} (IDTD) as
\begin{equation}
    D(A,B)
    =
    C(B)
    \left[
        1+\lambda d_{\to}(A,B)
    \right],
    \label{eq:latent_idtd}
\end{equation}
where \(C(B)\ge 0\) is the candidate capacity score,
\(d_{\to}(A,B)\in\mathbb R_{\ge 0}\) is a directed compatibility
discrepancy, and \(\lambda\ge 0\).
Lower IDTD indicates an easier transition toward the
candidate expert.

Because the factors in Eq.~\eqref{eq:latent_idtd} are not separately
observable, trajectory construction uses a rank-oriented operational score.
For each ordered pair, the detectors are frozen, and a collection of
scale-wise directional adaptors parameterized 
\begin{figure*}[t]
    \centering
    \includegraphics[width=\textwidth]{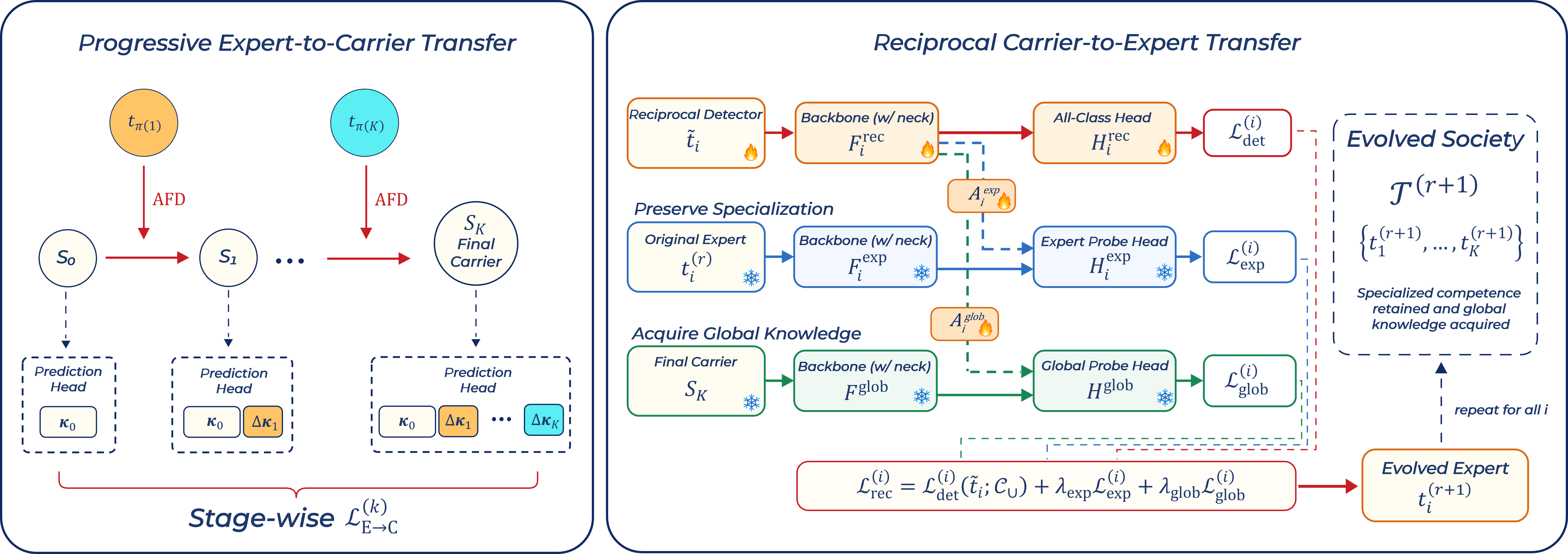}
    \caption{Progressive and reciprocal knowledge transfer in TGRD. Left: Along the planned trajectory, experts transfer knowledge via AFD to a carrier whose prediction head expands over accumulated categories, yielding $S_K$. Right: A reciprocal detector initialized from each expert is guided by the corresponding frozen original expert and the frozen $S_K$; the updated detectors form $\mathcal{T}^{(r+1)}$.}
    \label{fig:knowledge_transfer}
\end{figure*}
by
\(\theta_{A\to B}\) is fitted to align semantically matched,
standardized multi-scale representations of \(A\) with those of \(B\).
For an image \(x\),
\(\mathcal{L}_{\mathrm{align}}(x;\theta_{A\to B})\) denotes the
element-normalized sum of squared Frobenius residuals across the matched
semantic scales. The held-out residual defines

\begin{equation}
    \widehat{D}(A,B)
    =
    \frac{1}{|\mathcal{D}_{\mathrm{eval}}^{\cap}|}
    \sum_{x\in\mathcal{D}_{\mathrm{eval}}^{\cap}}
    \mathcal{L}_{\mathrm{align}}
    (x;\widehat{\theta}_{A\to B}),
    \label{eq:operational_idtd}
\end{equation}
where \(\widehat{\theta}_{A\to B}\) is obtained by minimizing the same
per-image alignment loss on \(\mathcal{D}_{\mathrm{fit}}^{\cap}\).
The detailed feature interface, standardization operation, and
scale-wise adaptor construction are provided in the Supplementary
Material under \emph{Operational IDTD Estimation}.

The score is directional, so generally
\(\widehat{D}(A,B)\ne\widehat{D}(B,A)\). At the
latent-to-operational surrogate layer, we assume anchor-wise order consistency:
for every fixed anchor \(A\), \(\widehat{D}(A,\cdot)\) preserves the candidate
ordering induced by \(D(A,\cdot)\).

\paragraph{Fixed-table greedy planning.}
Before carrier training, we precompute the directed table with entries
\(M_{0i}=\widehat{D}(S_0,t_i)\) and
\(M_{ji}=\widehat{D}(t_j,t_i)\) for \(j\ne i\). The table remains fixed
during carrier training. With
\(\mathcal{I}_0=\{1,\ldots,K\}\) and \(a_0=S_0\), the directed greedy
trajectory is
\begin{equation}
    \begin{aligned}
        \pi(k)
        &\in
        \operatorname*{arg\,min}_{i\in\mathcal{I}_{k-1}}
        \widehat{D}(a_{k-1},t_i),
        \\
        \mathcal{I}_k
        &=
        \mathcal{I}_{k-1}\setminus\{\pi(k)\},
        \qquad
        a_k=t_{\pi(k)}.
    \end{aligned}
    \label{eq:greedy_planner}
\end{equation}
Thus the first decision uses the \(S_0\) row and each later decision uses the
row of the most recently visited expert. 
The offline anchor \(a_{k-1}\) serves as a proxy for the actual carrier
\(S_{k-1}\). A sufficient latent-cost fidelity-and-margin condition under
which \(D(a_{k-1},\cdot)\) and \(D(S_{k-1},\cdot)\) select the same expert
is provided in the Supplementary Material under
\emph{Planning-Row Fidelity}. The implemented \(\widehat{D}\)-based planner
is linked to this latent selection through the separate anchor-wise
order-consistency condition above.

\subsection{Progressive and Reciprocal Knowledge Transfer}

Given \(\pi\), TGRD instantiates the SDL protocol
\(\Omega^{(r)}\) through forward expert-to-carrier consolidation
followed by reciprocal carrier-to-expert transfer, as illustrated
in Figure~\ref{fig:knowledge_transfer}.

\paragraph{Progressive expert-to-carrier transfer.}
Let \(\mathcal{D}_0\) be the initial carrier data and
\(\mathcal{D}_i\) the data associated with expert \(t_i\). At stage \(k\),
\(S_k\) is initialized from \(S_{k-1}\), its head is expanded according to
Eq.~\eqref{eq:progressive_categories}, and it is trained on
\begin{equation}
\mathcal{D}_{\mathrm{cum}}^{(k)}
:=
\mathcal{D}_0
\cup
\bigcup_{j=1}^{k}
\mathcal{D}_{\pi(j)} .
\label{eq:cumulative_data}
\end{equation}
whose annotations are exhaustive over \(\mathcal{K}_k\). 
Using the semantic-scale matching and standardization interface
underlying Eq.~(7), let \(\mathcal Q_k\) denote the set of matched
semantic-scale pairs between \(S_k\) and \(t_{\pi(k)}\). For each
\(q\in\mathcal Q_k\), let \(Z_q^{S_k}\) and
\(Z_q^{t_{\pi(k)}}\) denote the corresponding carrier and expert
representations, respectively. At each matched scale, a
stage-specific trainable directional adaptor maps the carrier
representation into the corresponding representation space of the
supervising expert. We refer to the resulting adaptor-mediated
feature transfer as Adaptive Feature Distillation (AFD); the adaptor
parameters are suppressed in Eq.~(10) for notational brevity.
\begin{equation}
    \mathcal{L}_{\mathrm{E\to C}}^{(k)}
    =
    \mathcal{L}_{\mathrm{det}}^{(k)}
    (S_k;\mathcal{K}_k)
    +
    \sum_{q\in\mathcal{Q}_k}
    \mathcal{L}_{\mathrm{AFD}}^{q}
    \!\left(
        Z_q^{S_k},
        Z_q^{t_{\pi(k)}}
    \right).
    \label{eq:forward_transfer}
\end{equation}
Each \(\mathcal L_{\mathrm{AFD}}^q\) is the corresponding scale-wise
component of the per-image, element-normalized alignment formulation
underlying Eq.~(7), instantiated with the stage-specific AFD adaptor.
The task loss supervises both newly introduced and
previously acquired categories, while AFD aligns the matched carrier
and expert representations. During stage \(k\), \(S_k\) and the AFD
adaptors are optimized, whereas the supervising expert
\(t_{\pi(k)}\) remains frozen. The fitted IDTD adaptors in
Eq.~\eqref{eq:operational_idtd} are used only for offline planning
and are not reused in carrier training. After all stages,
\(S_K\) supports \(\mathcal{C}_{\cup}\).

\paragraph{Reciprocal carrier-to-expert transfer.}
For every original expert \(t_i^{(r)}\), we initialize a trainable reciprocal
detector \(\widetilde{t}_i\) from \(t_i^{(r)}\). Its prediction head is
expanded to \(\mathcal{C}_{\cup}\): parameters for categories in
\(\mathcal{C}_i\) are retained by exact name matching, and the remaining
outputs are initialized. The resulting all-class head is denoted by
\(H_i^{\mathrm{rec}}\), and 
\(\widetilde t_i\) is trained on
\(\mathcal D_{\mathrm{cum}}^{(K)}\)
, whose annotations are exhaustive over
\(\mathcal{C}_{\cup}\).

The original expert and final carrier remain frozen and provide an expert
probe head \(H_i^{\mathrm{exp}}\) and a global probe head
\(H^{\mathrm{glob}}\), respectively. For a common input \(x\), let
\(F_i^{\mathrm{rec}}(x)\), \(F_i^{\mathrm{exp}}(x)\), and
\(F^{\mathrm{glob}}(x)\) denote their corresponding Backbone with Neck features.
Trainable adaptors \(A_i^{\mathrm{exp}}\) and \(A_i^{\mathrm{glob}}\) map
reciprocal features into the input spaces of the two frozen heads. 
Let \(\mathrm{KD}(z_s,z_t)\) denote the adopted
classification-logit distillation loss, with student-path and
reference logits as its first and second arguments, respectively.
The cross-head losses are
\begin{equation}
\begin{aligned}
\mathcal{L}_{\mathrm{exp}}^{(i)}
&=
\mathrm{KD}\!\Bigl(
    H_i^{\mathrm{exp}}\!\bigl(
        A_i^{\mathrm{exp}}(F_i^{\mathrm{rec}}(x))
    \bigr),
    H_i^{\mathrm{exp}}\!\bigl(
        F_i^{\mathrm{exp}}(x)
    \bigr)
\Bigr),
\\
\mathcal{L}_{\mathrm{glob}}^{(i)}
&=
\mathrm{KD}\!\Bigl(
    H^{\mathrm{glob}}\!\bigl(
        A_i^{\mathrm{glob}}(F_i^{\mathrm{rec}}(x))
    \bigr),
    H^{\mathrm{glob}}\!\bigl(
        F^{\mathrm{glob}}(x)
    \bigr)
\Bigr).
\end{aligned}
\label{eq:reciprocal_kd}
\end{equation}
Both terms distill only the classification logits of their corresponding
frozen probe heads. The reciprocal objective is
\begin{equation}
    \mathcal{L}_{\mathrm{rec}}^{(i)}
    =
    \mathcal{L}_{\mathrm{det}}^{(i)}
    (\widetilde{t}_i;\mathcal{C}_{\cup})
    +
    \lambda_{\mathrm{exp}}
    \mathcal{L}_{\mathrm{exp}}^{(i)}
    +
    \lambda_{\mathrm{glob}}
    \mathcal{L}_{\mathrm{glob}}^{(i)}.
    \label{eq:reciprocal_objective}
\end{equation}
Here \(\lambda_{\mathrm{exp}},\lambda_{\mathrm{glob}}\geq 0\)
weight the expert-specific and society-wide guidance terms,
respectively.
Only \(\widetilde{t}_i\), \(A_i^{\mathrm{exp}}\), and
\(A_i^{\mathrm{glob}}\) are optimized. After training,
\(t_i^{(r+1)}=\widetilde{t}_i\), so replacing every society member completes
the SDL update in Eq.~\eqref{eq:sdl_update}. At inference, only the updated
detector and its all-class head \(H_i^{\mathrm{rec}}\) are used.

The following section provides a conditional proxy-certificate
comparison between the progressive expert-to-carrier construction
and an abstract single aggregated-target alternative.

\section{Theoretical Analysis}
\label{sec:theory}

We analyze the progressive expert-to-carrier phase under a
target-wise certificate abstraction. Fix a realized complete
trajectory \(\pi\), and let \(g_i\) be the expert-specific transfer
target induced by \(t_i\). We use \(g_{\mathrm{agg}}\) only as an
abstract single aggregated-target comparator. The analysis concerns
the expert-dependent transfer component; cumulative detection
supervision remains part of the actual stage objective.

\textbf{Ambient setup.}
Let \(\mathcal F_S^\cup\) be the fixed hypothesis class associated
with the final union-category carrier architecture, and let
\(\mathcal F_S^{(k)}\) be the physically instantiated class at stage
\(k\). 
For the fixed trajectory \(\pi\), let
\(P_k:\mathcal F_S^\cup\to\mathcal F_S^{(k)}\) denote the
name-aligned restriction of the output blocks ordered by
\(\boldsymbol{\kappa}_K\) to the prefix
\(\boldsymbol{\kappa}_k\) constructed in Eq.~(5), and assume
\(P_k(\mathcal F_S^\cup)=\mathcal F_S^{(k)}\).
Identifying the physical carrier with the predictor it realizes,
\(S_k\in\mathcal F_S^{(k)}\); hence it admits an analytical ambient
extension \(\widehat f_k\in\mathcal F_S^\cup\) satisfying
\(P_k(\widehat f_k)=S_k\); 
this does not require a union-category head
to be physically instantiated before the final stage. Let
\(\widehat f_{\mathrm{agg}}\in\mathcal F_S^\cup\) denote the learner
for the aggregated comparator. The complete ordered-head construction
is given in the Supplementary Material.

For \(g\in\mathcal G:=\{g_1,\ldots,g_K,g_{\mathrm{agg}}\}\), let
\(R_g\) be its target-specific population risk, with optima
\(R_g^\star\) over a reference class
\(\mathcal F\supseteq\mathcal F_S^\cup\) and \(R_{g,S}^\star\) over
\(\mathcal F_S^\cup\). Write \(R_i:=R_{g_i}\) and
\(R_{\mathrm{agg}}:=R_{g_{\mathrm{agg}}}\), and define
\begin{equation}
\begin{aligned}
\operatorname{Est}_k(n)
&:= R_{\pi(k)}(\widehat f_k)
   - R_{\pi(k),S}^{\star},\\
\epsilon_i
&:= R_{i,S}^{\star}-R_i^{\star},
   \qquad i=1,\ldots,K,\\
\operatorname{Est}_{\mathrm{agg}}(n)
&:= R_{\mathrm{agg}}(\widehat f_{\mathrm{agg}})
   - R_{\mathrm{agg},S}^{\star},\\
\epsilon_{\mathrm{agg}}
&:= R_{\mathrm{agg},S}^{\star}
   - R_{\mathrm{agg}}^{\star}.
\end{aligned}
\label{eq:theory_decomposition_terms}
\end{equation}
For each stage \(k\) with \(i=\pi(k)\), assume that the restriction
of \(R_i\) to \(\mathcal F_S^\cup\) factors through \(P_k\);
equivalently, for all \(f,f'\in\mathcal F_S^\cup\),
\[
P_k(f)=P_k(f')
\quad\Longrightarrow\quad
R_i(f)=R_i(f').
\]
Consequently, \(P_k(\widehat f_k)=S_k\) makes
\(R_i(\widehat f_k)\) independent of the chosen ambient extension.
The approximation burdens \(\epsilon_i\) and
\(\epsilon_{\mathrm{agg}}\) are nonnegative because
\(\mathcal F_S^\cup\subseteq\mathcal F\).
Here \(n\ge1\) is a common certificate-evaluation budget. We suppress
the dependence of \(\widehat f_k\) on \(n\) and the fixed trajectory
\(\pi\), and that of \(\widehat f_{\mathrm{agg}}\) on \(n\). This
abstraction does not assert equal total training data or compute.

\textbf{A1 (simultaneous target-wise certificates).}
Assume that there are constants \(A_i,A_{\mathrm{agg}}>0\), a shared
\(\mathcal C_S:(0,1)\to(0,\infty)\), and designated exponents
\(\alpha_i,\alpha_{\mathrm{agg}}\in[1/2,1]\) such that, for every
\(n\ge1\) and \(\delta\in(0,1)\), with probability at least
\(1-\delta\), simultaneously for all \(k\),
\begin{equation}
\begin{aligned}
\operatorname{Est}_k(n)
 &\le A_{\pi(k)}\mathcal C_S(\delta)n^{-\alpha_{\pi(k)}},\\
\operatorname{Est}_{\mathrm{agg}}(n)
 &\le A_{\mathrm{agg}}\mathcal C_S(\delta)
       n^{-\alpha_{\mathrm{agg}}}
\end{aligned}
\label{eq:theory_a1}
\end{equation}
The event is joint across all displayed targets; for a data-dependent
planner, its validity is understood uniformly over the admissible
complete trajectories. 

\textbf{A2 (proxy-to-exponent monotonicity).}
Using the latent IDTD \(D\) from the Method, define the same-anchor
expert proxies and the aggregated-target proxy by
\begin{equation}
\begin{aligned}
\overline D_i&:=D(S_0,t_i),\qquad i=1,\ldots,K,\\
\overline D_{\mathrm{agg}}
&:=\max_i\overline D_i+\Gamma(\mathcal T),
\quad \Gamma(\mathcal T)\ge0.
\end{aligned}
\label{eq:theory_proxies}
\end{equation}
Set \(\overline D(g_i)=\overline D_i\) and
\(\overline D(g_{\mathrm{agg}})=\overline D_{\mathrm{agg}}\).
Write \(\alpha(g_i)=\alpha_i\) and
\(\alpha(g_{\mathrm{agg}})=\alpha_{\mathrm{agg}}\). For
\(g_a,g_b\in\mathcal G\), assume
\begin{equation}
\overline D(g_a)\le\overline D(g_b)
\quad\Longrightarrow\quad
\alpha(g_a)\ge\alpha(g_b).
\label{eq:theory_a2}
\end{equation}
A2 is an explicit property of the selected certificate family, not a
universal learning law. Since every
\(\overline D_i\le\overline D_{\mathrm{agg}}\), it gives
\begin{equation}
\alpha_{\mathrm{prog}}
:=\min_k\alpha_{\pi(k)}
=\min_i\alpha_i
\ge\alpha_{\mathrm{agg}}.
\label{eq:theory_exponent_order}
\end{equation}
Thus the comparison applies to any complete expert order; it does not
distinguish or optimize complete permutations.

Let
\begin{equation}
\begin{gathered}
A_{\mathrm{prog}}:=\max_i A_i,\qquad
\epsilon_{\mathrm{prog}}:=\sum_{i=1}^K\epsilon_i,\\
B_{\mathrm{prog}}(n)
:=A_{\mathrm{prog}}K\mathcal C_S(\delta)
  n^{-\alpha_{\mathrm{prog}}}+\epsilon_{\mathrm{prog}},\\
B_{\mathrm{agg}}(n)
:=A_{\mathrm{agg}}\mathcal C_S(\delta)
  n^{-\alpha_{\mathrm{agg}}}+\epsilon_{\mathrm{agg}} .
\end{gathered}
\label{eq:theory_certificates}
\end{equation}
On the event in A1, these respectively upper-bound the constructed
sum of the \(K\) target-specific progressive excess risks and the
aggregated-target excess risk.

\textbf{Theorem 1 (conditional proxy-certificate comparison).}
Fix \(n\ge1\) and \(\delta\in(0,1)\). Under the ambient setup and A1--A2, let
\(\Delta\alpha:=\alpha_{\mathrm{prog}}-\alpha_{\mathrm{agg}}\ge0\).
If
\begin{equation}
A_{\mathrm{prog}}K
\le A_{\mathrm{agg}}n^{\Delta\alpha},
\qquad
\epsilon_{\mathrm{prog}}\le\epsilon_{\mathrm{agg}},
\label{eq:theory_comparison_conditions}
\end{equation}
then
\begin{equation}
B_{\mathrm{prog}}(n)\le B_{\mathrm{agg}}(n).
\label{eq:theory_certificate_comparison}
\end{equation}

\noindent\emph{Proof.}
Equation~\eqref{eq:theory_proxies} and A2 give
Eq.~\eqref{eq:theory_exponent_order}. Summing A1 over the progressive
stages and using \(n\ge1\) bounds their estimation terms by
\(A_{\mathrm{prog}}K\mathcal C_S(\delta)
n^{-\alpha_{\mathrm{prog}}}\). The first condition in
Eq.~\eqref{eq:theory_comparison_conditions} makes this no larger than
the aggregated estimation certificate; adding the second condition
proves the result. \(\square\)

This result compares proxy certificates for different target-specific risks rather than actual detection error under a common risk; further details are provided in the Supplementary Material.

\section{Experiments}

\subsection{Experimental Setup}

\paragraph{Dataset and detector society.}
We conduct experiments on MS COCO 2017~\cite{lin2014microsoft} using \texttt{train2017} for
training and \texttt{val2017} for evaluation. 
We set $K=4$ and form the society with four independently trained heterogeneous experts: RetinaNet~\cite{lin2017focal}, FCOS~\cite{tian2019fcos}, Faster R-CNN~\cite{ren2015faster}, and GFL~\cite{li2020generalized}; their assignments are reported in Table~\ref{tab:fixed_table_planning}. The initial carrier support
satisfies $|\mathcal C_0|=40$, and each expert support $\mathcal C_i$
extends $\mathcal C_0$ with 10 expert-specific categories. The four
expert-specific increments $\mathcal C_i\setminus\mathcal C_0$ are
pairwise disjoint, yielding $\mathcal C_\cap=\mathcal C_0$ and
$|\mathcal C_\cup|=80$. We instantiate $S_0$ as either RetinaNet or
Faster R-CNN and train it on $\mathcal D_0$ over $\mathcal C_0$.

\begin{table}[t]
\centering
\small
\setlength{\tabcolsep}{3.5pt}
\renewcommand{\arraystretch}{1.02}

\begin{tabular}{@{}lcccc@{}}
\toprule
\multicolumn{5}{c}{\textbf{(a) Operational IDTD scores}
$\widehat D(a_{k-1},t_i)$} \\
& \multicolumn{4}{c}{Candidate next expert $t_i$} \\
\cmidrule(lr){2-5}
Planning anchor $a_{k-1}$
& $t_1$ & $t_2$ & $t_3$ & $t_4$ \\
\midrule
$S_0$ (RetinaNet)
& 0.5463 & 1.2115 & 0.5588 & 0.6411 \\
$S_0$ (Faster R-CNN)
& 0.5148 & 1.1658 & 0.2318 & 0.5473 \\
\midrule
$t_1$ (RetinaNet)
& -- & 1.1792 & 0.5514 & 0.4862 \\
$t_2$ (FCOS)
& 1.1392 & -- & 1.0478 & 1.1748 \\
$t_3$ (Faster R-CNN)
& 0.5075 & 1.1638 & -- & 0.5403 \\
$t_4$ (GFL)
& 0.4639 & 1.1745 & 0.5478 & -- \\
\midrule
\multicolumn{5}{c}{\textbf{(b) Greedy planning trajectories}} \\
Initial carrier
& \multicolumn{4}{l}{Greedy planning trajectory} \\
\cmidrule(lr){1-1}
\cmidrule(lr){2-5}
RetinaNet
& \multicolumn{4}{l}{$S_0\rightarrow t_1\rightarrow t_4
\rightarrow t_3\rightarrow t_2$} \\
Faster R-CNN
& \multicolumn{4}{l}{$S_0\rightarrow t_3\rightarrow t_1
\rightarrow t_4\rightarrow t_2$} \\
\bottomrule
\end{tabular}
\caption{Operational IDTD scores and fixed-table greedy trajectories
under two carrier initializations. Detector architectures are given
in parentheses; all configurations use R50--FPN.}
\label{tab:fixed_table_planning}
\end{table}

\paragraph{Protocol and evaluation.}
For each ordered pair $(A,B)$ of planning anchor $A$ and candidate
expert $B$, we compute $\widehat D(A,B)$ over $\mathcal C_\cap$ by
fitting scale-wise directional $1\times1$ adaptors on
$\mathcal D_{\mathrm{fit}}^\cap$ and evaluating the alignment
residual on the disjoint $\mathcal D_{\mathrm{eval}}^\cap$. The
resulting score table is fixed before carrier training and determines
the greedy permutation $\pi$. At stage $k$, $S_{k-1}$ is updated to
$S_k$ on $\mathcal D_{\mathrm{cum}}^{(k)}$, whose annotations are
exhaustive over $\mathcal K_k$. After $K$ stages, the frozen original
expert $t_i$ and final carrier $S_K$ jointly guide each reciprocal
detector $\widetilde t_i$ on $\mathcal D_{\mathrm{cum}}^{(K)}$. We
evaluate one socialization round using standard COCO bounding-box AP.


To our knowledge, no prior detection protocol jointly considers complementary heterogeneous experts, progressive union-category carrier construction, and reciprocal expert updates; fixed-category multi-teacher distillation and class-incremental detection are therefore not protocol-matched. We instead use Avg-FPN KD, which simultaneously distills the four experts' averaged FPN features into $S_0$ under the same society, data, annotations, evaluation protocol, and matched 48-epoch budget.

\subsection{Experimental Results}

\paragraph{IDTD-guided planning.}
As shown in Table~\ref{tab:fixed_table_planning}, RetinaNet first
selects $t_1$ ($0.5463$), whereas Faster R-CNN first selects $t_3$
($0.2318$). The ensuing row-wise greedy choices produce
$t_1\!\rightarrow\!t_4\!\rightarrow\!t_3\!\rightarrow\!t_2$ and
$t_3\!\rightarrow\!t_1\!\rightarrow\!t_4\!\rightarrow\!t_2$,
respectively, showing that the operational order depends on the
initial carrier. In both cases, whenever $t_2$ remains eligible, it
has the largest score among the remaining candidates and is selected
last.

\begin{table*}[!t]
\centering
\small
\setlength{\tabcolsep}{5.0pt}
\renewcommand{\arraystretch}{1.02}

\begin{tabular}{@{}lcccccc@{}}
\toprule
Carrier state / control
& $AP$
& $AP_{\mathcal C_0}$
& $AP_{\Delta\kappa_1}$
& $AP_{\Delta\kappa_2}$
& $AP_{\Delta\kappa_3}$
& $AP_{\Delta\kappa_4}$ \\
\midrule
\multicolumn{7}{@{}l}{\textbf{(a) RetinaNet as initial carrier $S_0$}} \\
$S_0$
& 37.3 & 37.3 & -- & -- & -- & -- \\
$S_1\,[t_1]$
& 35.9 & 39.2 & 22.8 & -- & -- & -- \\
$S_2\,[t_1,t_4]$
& 35.1 & 40.5 & 22.6 & 26.4 & -- & -- \\
$S_3\,[t_1,t_4,t_3]$
& 34.5 & 39.6 & 19.8 & 39.4 & 24.1 & -- \\
$S_4\,[t_1,t_4,t_3,t_2]$
& 32.8 & 39.5 & 21.0 & 22.2 & 38.7 & 22.0 \\
Simultaneous aggregation (Avg-FPN KD)
& 30.2 & 37.9 & 14.5 & 18.5 & 37.2 & 19.3 \\
\midrule
\multicolumn{7}{@{}l}{\textbf{(b) Faster R-CNN as initial carrier $S_0$}} \\
$S_0$
& 40.0 & 40.0 & -- & -- & -- & -- \\
$S_1\,[t_3]$
& 38.8 & 39.0 & 38.0 & -- & -- & -- \\
$S_2\,[t_3,t_1]$
& 36.3 & 40.0 & 39.3 & 18.5 & -- & -- \\
$S_3\,[t_3,t_1,t_4]$
& 34.5 & 39.7 & 39.2 & 20.4 & 23.0 & -- \\
$S_4\,[t_3,t_1,t_4,t_2]$
& 33.2 & 39.1 & 39.1 & 21.3 & 25.5 & 23.8 \\
Simultaneous aggregation (Avg-FPN KD)
& 30.6 & 39.6 & 36.5 & 13.5 & 15.3 & 20.9 \\
\bottomrule
\end{tabular}
\caption{Stage-wise carrier performance from $S_0$ along two
fixed-table trajectories versus simultaneous aggregation (Avg-FPN KD).
Overall AP is evaluated on $\mathcal K_k$ for $S_k$ and on
$\mathcal C_\cup$ for the control, with
$\mathcal K_0=\mathcal C_0$. $AP_{\Delta\kappa_j}$ denotes AP over
the categories introduced by $t_{\pi(j)}$ that are absent from
$\mathcal K_{j-1}$; control columns follow the same
trajectory-specific order. Brackets list the incorporated experts.}
\label{tab:progressive_vs_simultaneous}
\end{table*}


\begin{table*}[t]
\centering
\small
\begin{tabular}{@{}l*{8}{c}@{}}
\toprule
&
\multicolumn{3}{c}{Initial support $\mathcal{C}_i$}
&
\multicolumn{3}{c}{Expert-specific support
$\mathcal{C}_i\setminus\mathcal{C}_0$}
&
\multicolumn{2}{c}{Knowledge expansion} \\
\cmidrule(lr){2-4}
\cmidrule(lr){5-7}
\cmidrule(lr){8-9}

Expert
& Original
& Reciprocal
& $\Delta$
& Original
& Reciprocal
& $\Delta$
& $AP_{\mathcal{C}_{\cup}\setminus\mathcal{C}_i}$
& $AP_{\mathcal{C}_{\cup}}$ \\
\midrule

\multicolumn{9}{@{}l}{%
\textbf{(a)} Final RetinaNet carrier $S_4$} \\

$t_1$
& 35.5 & 35.9 & $+0.4$
& 22.5 & 22.3 & $-0.2$
& 25.8 & 32.1 \\

$t_2$
& 40.0 & 39.2 & $-0.8$
& 31.7 & 30.6 & $-1.1$
& 25.6 & 34.1 \\

$t_3$
& 39.8 & 40.1 & $+0.3$
& 40.2 & 40.7 & $+0.5$
& 23.2 & 33.7 \\

$t_4$
& 38.8 & 38.9 & $+0.1$
& 26.5 & 27.8 & $+1.3$
& 28.4 & 35.0 \\

\midrule

\multicolumn{9}{@{}l}{%
\textbf{(b)} Final Faster R-CNN carrier $S_4$} \\

$t_1$
& 35.5 & 35.5 & $0.0$
& 22.5 & 22.4 & $-0.1$
& 24.2 & 31.2 \\

$t_2$
& 40.0 & 39.6 & $-0.4$
& 31.7 & 30.7 & $-1.0$
& 24.5 & 33.9 \\

$t_3$
& 39.8 & 39.9 & $+0.1$
& 40.2 & 40.6 & $+0.4$
& 20.8 & 32.8 \\

$t_4$
& 38.8 & 38.7 & $-0.1$
& 26.5 & 26.5 & $ 0.0$
& 26.6 & 34.1 \\

\bottomrule
\end{tabular}

\caption{Expert retention and knowledge expansion after reciprocal
updates jointly guided by the original experts and final
(a) RetinaNet or (b) Faster R-CNN carriers. $\Delta$ denotes the
reciprocal-minus-original AP change.}
\label{tab:reciprocal_evolution}
\end{table*}

\paragraph{Progressive expert-to-carrier transfer.}
Table~\ref{tab:progressive_vs_simultaneous} follows each carrier from
$S_0$ to $S_4$. Because intermediate overall AP is evaluated on the
growing support $\mathcal K_k$, cross-stage values are not directly
comparable as a conventional performance trend. At $S_4$,
$\mathcal K_4=\mathcal C_\cup$, enabling a like-for-like comparison
with the simultaneous-aggregation control. The final RetinaNet and
Faster R-CNN carriers obtain $32.8$ and $33.2$ AP, respectively,
exceeding the corresponding Avg-FPN KD controls by $2.6$ AP in both
settings. All four increment-wise comparisons also favor the final
carrier: the gains are $1.5$--$6.5$ AP for RetinaNet and
$2.6$--$10.2$ AP for Faster R-CNN, with averages of $3.6$ and
$5.9$ AP.

The $S_0$ rows permit a direct check of initial-support
retention. For RetinaNet, $AP_{\mathcal C_0}$ changes from $37.3$ at
$S_0$ to $39.5$ at $S_4$, compared with $37.9$ for the control. For
Faster R-CNN, the final value of $39.1$ is within $0.9$ AP of $S_0$
and $0.5$ AP of the control. Because both procedures use the same
total 48-epoch budget, these gains are not explained by a longer total
training schedule. Thus, under both initializations, progressive
transfer outperforms the matched simultaneous control at the union and
increment levels while largely preserving the initial support.

\paragraph{Reciprocal carrier-to-expert transfer.}
Table~\ref{tab:reciprocal_evolution} evaluates the reciprocal detectors
after joint guidance from their frozen original experts and the final
carriers. Across all eight expert--carrier pairs, the reciprocal
detectors reach $20.8$--$28.4$ AP on the previously unsupported
categories $\mathcal C_\cup\setminus\mathcal C_i$ and
$31.2$--$35.0$ AP over the full union support. At the same time, the
changes on the original supports $\mathcal C_i$ remain between
$-0.8$ and $+0.4$ AP, with a mean absolute change of only $0.28$ AP.
On the expert-specific supports
$\mathcal C_i\setminus\mathcal C_0$, every reciprocal detector remains
within $1.3$ AP of its original expert, with a mean absolute change of
$0.58$ AP. These results show that the complete reciprocal procedure
expands each expert to society-wide category coverage with only
limited changes on its original and expert-specific supports. Together
with the carrier results in
Table~\ref{tab:progressive_vs_simultaneous}, the observed pattern is
consistent with the intended bidirectional socialization outcome under
both carrier initializations.

\section{Conclusion}

We formulated Socialized Detector Learning (SDL) to treat complementary knowledge as a resource of a heterogeneous detector society and to evolve both the society and its members through exchange. TGRD instantiates one socialization round through directed IDTD-based trajectory planning, progressive union-category carrier construction, and reciprocal carrier-to-expert transfer. Under stated assumptions, our conditional target-specific proxy-certificate analysis shows that the progressive certificate is no larger than an aggregated-target counterpart. Across two carrier initializations on MS COCO, final carriers outperform epoch-matched simultaneous aggregation by 2.6 AP, while reciprocal detectors attain 20.8--28.4 AP on previously unsupported categories and remain within 1.3 AP of original expert-specific performance. These results support order-aware consolidation and reciprocal transfer as a viable path to broader society-wide coverage with retained specialization, moving detector learning beyond one-way student enhancement toward collective detector-society evolution.

\bibliography{aaai2027}
\clearpage
\appendix

\section*{Supplement Overview}
This supplement contains four parts that expand the technical details deferred
from the main paper:
\begin{enumerate}
  \setlength{\itemsep}{0pt}
  \setlength{\parskip}{0pt}
  \item \textbf{Operational IDTD Estimation} defines the common feature
  interface, spatial standardization, directional adaptors, and held-out score.
  \item \textbf{Ordered-Head Construction} specifies the union-category order,
  stage-wise head expansion, and exact category-name alignment.
  \item \textbf{Conditional Proxy-Certificate Details} states the certificate
  assumptions and proves the finite-sample comparison used in the main paper.
  \item \textbf{Planning-Row Fidelity} gives a sufficient fidelity-and-margin
  condition under which offline planning preserves the carrier-aware choice.
\end{enumerate}
The definitions of Socialized Detector Learning (SDL), the overall TGRD
procedure, and the progressive and reciprocal training objectives are not
repeated here.

\appendix
\renewcommand{\theequation}{\arabic{equation}}
\setcounter{equation}{0}
\allowdisplaybreaks

\section{Operational IDTD Estimation}
\label{sec:operational-idtd}

This section specifies the feature interface, standardization operation, and
scale-wise directional adaptors used to compute the operational score
$\widehat D(A,B)$. For every ordered pair $(A,B)$, the two detectors are frozen
and kept in evaluation mode. Adaptor fitting and held-out evaluation use the
same nonempty, finite, and disjoint probe-image sets
$\mathcal D_{\mathrm{fit}}^{\cap}$ and
$\mathcal D_{\mathrm{eval}}^{\cap}$. The sets are constructed by the same
sampling rule over
$\mathcal C_{\cap}=\mathcal C_0\cap\bigcap_{i=1}^{K}\mathcal C_i\neq\varnothing$,
where $\mathcal C_0$ and $\mathcal C_i$ are the initial carrier and expert
category sets. Every
sample contains at least one annotated instance from $\mathcal C_{\cap}$.

\subsection{Generalized feature interface}

Each detector exposes intermediate spatial representations through a common
semantic-scale interface. For an ordered pair $(A,B)$, let
$\mathcal Q_{A,B}$ index representations selected at corresponding semantic
scales. For $q\in\mathcal Q_{A,B}$ and $M\in\{A,B\}$, let
$F_q^M(x)$ denote the representation of detector $M$ selected at scale $q$.
A fixed reshape or resampling operator $R_q^M$ maps this representation to the
canonical spatial resolution of the matched scale:
\begin{equation}
  Z_q^M(x) \coloneqq R_q^M\!\left(F_q^M(x)\right).
  \label{eq:supp-feature-interface}
\end{equation}
The interface therefore permits heterogeneous backbones and feature pyramids,
provided that spatial representations with compatible semantic scales can be
selected and resampled.

For a tensor $F\in\mathbb R^{N_b\times C\times H\times W}$, where $N_b$ is
the batch size, we apply per-sample, per-channel spatial standardization:
\begin{align}
  \mu_{b,c}(F)
  &\coloneqq \frac{1}{HW}\sum_{h=1}^{H}\sum_{w=1}^{W}F_{b,c,h,w},
  \label{eq:supp-mean}\\
  \sigma_{b,c}^{2}(F)
  &\coloneqq \frac{1}{HW}\sum_{h=1}^{H}\sum_{w=1}^{W}
  \bigl(F_{b,c,h,w}-\mu_{b,c}(F)\bigr)^2,
  \label{eq:supp-var}\\
  \mathcal N(F)_{b,c,h,w}
  &\coloneqq
  \frac{F_{b,c,h,w}-\mu_{b,c}(F)}
       {\sqrt{\sigma_{b,c}^{2}(F)+\varepsilon}},
  \qquad \varepsilon>0.
  \label{eq:supp-standardization}
\end{align}

For every ordered pair and every matched scale, a separate directional
$1\times1$ convolutional adaptor $a_{A\to B,q}$ maps the channel dimension of
$A$ to that of $B$. Define
\begin{equation}
  \begin{aligned}
    \widetilde Z_q^{A\to B}(x)
    &\coloneqq
    \mathcal N\!\left(a_{A\to B,q}\!\left(Z_q^A(x)\right)\right),\\
    \widetilde Z_q^B(x)
    &\coloneqq \mathcal N\!\left(Z_q^B(x)\right).
  \end{aligned}
  \label{eq:supp-adapted-features}
\end{equation}
The per-image, element-normalized alignment loss is
\begin{equation}
  \mathcal L_{\mathrm{align}}(x;\theta_{A\to B})
  \coloneqq
  \frac{
    \displaystyle\sum_{q\in\mathcal Q_{A,B}}
    \left\lVert
      \widetilde Z_q^{A\to B}(x)-\widetilde Z_q^B(x)
    \right\rVert_F^2
  }{
    \displaystyle\sum_{q\in\mathcal Q_{A,B}}
    \operatorname{numel}\!\left(Z_q^B(x)\right)
  },
  \label{eq:supp-alignment-loss}
\end{equation}
where $\theta_{A\to B}$ collects the adaptors for all matched scales. Thus,
every normalized feature element receives equal weight across scales.

\subsection{Adaptor fitting and held-out scoring}

Only the adaptor parameters are optimized on
$\mathcal D_{\mathrm{fit}}^{\cap}$:
\begin{equation}
  \widehat\theta_{A\to B}
  \in
  \operatorname*{arg\,min}_{\theta_{A\to B}}
  \frac{1}{\lvert\mathcal D_{\mathrm{fit}}^{\cap}\rvert}
  \sum_{x\in\mathcal D_{\mathrm{fit}}^{\cap}}
  \mathcal L_{\mathrm{align}}(x;\theta_{A\to B}).
  \label{eq:supp-adaptor-fit}
\end{equation}
All ordered pairs use the same adaptor architecture, optimization schedule,
and training budget. The operational IDTD score is the held-out residual
\begin{equation}
  \widehat D(A,B)
  \coloneqq
  \frac{1}{\lvert\mathcal D_{\mathrm{eval}}^{\cap}\rvert}
  \sum_{x\in\mathcal D_{\mathrm{eval}}^{\cap}}
  \mathcal L_{\mathrm{align}}
  \bigl(x;\widehat\theta_{A\to B}\bigr).
  \label{eq:supp-operational-idtd}
\end{equation}
The score is directional because the adaptor maps the representation space of
$A$ into that of $B$; in general,
$\widehat D(A,B)\neq\widehat D(B,A)$.

The latent composite difficulty in the main paper is
\begin{equation}
  D(A,B)=C(B)\bigl[1+\lambda d_{\to}(A,B)\bigr].
  \label{eq:supp-latent-idtd}
\end{equation}
Here $C(B)\geq0$ is the capacity score of candidate $B$,
$d_{\to}(A,B)\geq0$ is a directed compatibility discrepancy, and
$\lambda\geq0$ is its weight.
Its factors are not separately observable. Equation
\eqref{eq:supp-operational-idtd} is therefore a rank-oriented operational
surrogate for the composite effect, rather than a plug-in estimator of
$C(B)$, $\lambda$, and $d_{\to}(A,B)$. The planner requires only
anchor-wise order consistency: for a fixed anchor $A$ and candidate experts
$t_i,t_j$, a sufficiently separated latent ordering should be preserved by
the empirical row. The fitted IDTD adaptors are used only to construct this
offline table and are not reused by Adaptive Feature Distillation (AFD) during
carrier training.

\section{Ordered-Head Construction}
\label{sec:ordered-head}

Let $\mathcal C_0$ be the category set initially supported by the carrier
$S_0$, and let $\mathcal C_i$ be the category set of expert $t_i$. The final
union vocabulary is
\begin{equation}
  \mathcal C_{\cup}
  \coloneqq
  \mathcal C_0\cup\bigcup_{i=1}^{K}\mathcal C_i.
  \label{eq:supp-union-vocabulary}
\end{equation}
Let $c_0=(c_{0,1},\ldots,c_{0,\lvert\mathcal C_0\rvert})$ and
$c_i=(c_{i,1},\ldots,c_{i,\lvert\mathcal C_i\rvert})$ be fixed ordered
category-name lists. Category names are unique within each list, and shared
categories use the same canonical name across detectors.

For the expert trajectory $\pi$, initialize
$\kappa_0=c_0$ and $\mathcal K_0=\mathcal C_0$. At stage $k$, define
\begin{align}
  \Delta\mathcal C_k
  &\coloneqq \mathcal C_{\pi(k)}\setminus\mathcal K_{k-1},
  \label{eq:supp-new-category-set}\\
  \Delta\kappa_k
  &\coloneqq
  \bigl[c\in c_{\pi(k)}\mid c\in\Delta\mathcal C_k\bigr],
  \label{eq:supp-new-category-list}\\
  \kappa_k
  &\coloneqq \kappa_{k-1}\mathbin{\Vert}\Delta\kappa_k,
  \qquad
  \mathcal K_k\coloneqq\operatorname{set}(\kappa_k),
  \label{eq:supp-ordered-extension}
\end{align}
where brackets denote an order-preserving subsequence and $\Vert$ denotes
ordered concatenation. Consequently, $\kappa_{k-1}$ is a prefix of
$\kappa_k$ and $\mathcal K_K=\mathcal C_{\cup}$.

During the expansion from stage $k-1$ to stage $k$, the class-specific output
parameter blocks associated with the prefix $\kappa_{k-1}$ are copied into
the same prefix positions, while only the blocks associated with
$\Delta\kappa_k$ are initialized. Expert and carrier outputs are aligned by
exact category-name matching. In particular, for a local expert category
$c_{i,j}\in\mathcal K_k$, its carrier position is
\begin{equation}
  \rho_{i\to k}(j)
  \coloneqq \operatorname{pos}_{\kappa_k}(c_{i,j}).
  \label{eq:supp-index-map}
\end{equation}

To formalize the relationship between the progressively instantiated heads
and the final union-category architecture, let $L_{\cup}$ be a fixed canonical
ambient ordering of $\mathcal C_{\cup}$. Let
$\mathcal F_S^{\cup}$ be the fixed ambient hypothesis class whose
category-output blocks are indexed by $L_{\cup}$, and let
$\mathcal F_S^{(k)}$ be the physically instantiated class whose blocks are
indexed by $\kappa_k$. Define $P_k$ as the name-aligned
selection-and-reindexing operator from $L_{\cup}$ to $\kappa_k$; it acts as
the identity on all non-category output components. We assume
head-extension consistency:
\begin{equation}
  P_k\mathcal F_S^{\cup}
  \coloneqq
  \{P_k\circ f:f\in\mathcal F_S^{\cup}\}
  =\mathcal F_S^{(k)},
  \qquad k=0,\ldots,K.
  \label{eq:supp-head-extension-consistency}
\end{equation}
Thus every physical carrier $S_k\in\mathcal F_S^{(k)}$ admits an analytical
ambient extension $\widehat f_k\in\mathcal F_S^{\cup}$ satisfying
$P_k\circ\widehat f_k=S_k$. This construction does not require a physical
union-category head before the final stage.

For a transfer target $g$ whose supervised category coordinates
$\mathcal C(g)$ satisfy $\mathcal C(g)\subseteq\mathcal K_k$, let
$P_g^{\cup}$ select the corresponding named coordinates from the ambient
order, and let $P_g^{(k)}$ select them from the physical order $\kappa_k$.
Exact name alignment gives
\begin{equation}
  P_g^{\cup}=P_g^{(k)}\circ P_k.
  \label{eq:supp-target-selector-consistency}
\end{equation}
Therefore, the target-specific risk of an ambient extension depends only on
the physical stage predictor it represents.

\section{Conditional Proxy-Certificate Details}
\label{sec:certificate-details}

This section expands the conditional comparison stated in the main paper. It
concerns the progressive expert-to-carrier phase and an abstract
single-aggregated-target comparator. It does not analyze the reciprocal phase.

\subsection{Target-specific risks and certificate assumptions}

Let the expert society be
$\mathcal T=\{t_1,\ldots,t_K\}$. Define
\begin{equation*}
  \mathcal G=\{g_1,\ldots,g_K,g_{\mathrm{agg}}\},
\end{equation*}
where $g_i$ is the
expert-specific transfer target induced by $t_i$ and
$g_{\mathrm{agg}}$ is the abstract aggregated target. For a nonnegative
target-level loss $\ell$, define
\begin{equation}
  R_g(f)
  \coloneqq
  \mathbb E\!\left[
    \ell\!\left(P_g^{\cup}f(X),g(X)\right)
  \right].
  \label{eq:supp-target-risk}
\end{equation}
Let $\mathcal F\supseteq\mathcal F_S^{\cup}$ be a reference function class,
and define
\begin{equation}
  R_g^{\star}\coloneqq\inf_{f\in\mathcal F}R_g(f),
  \qquad
  R_{g,S}^{\star}\coloneqq
  \inf_{f\in\mathcal F_S^{\cup}}R_g(f).
  \label{eq:supp-risk-optima}
\end{equation}
For each target learner $\widehat f_g\in\mathcal F_S^{\cup}$ and a common
effective certificate-evaluation budget $n\geq1$, define
\begin{equation}
  \operatorname{Est}_g(n)
  \coloneqq R_g(\widehat f_g)-R_{g,S}^{\star},
  \qquad
  \epsilon(g)
  \coloneqq R_{g,S}^{\star}-R_g^{\star}.
  \label{eq:supp-estimation-approximation}
\end{equation}
Each target-specific excess risk then has the exact decomposition
\begin{equation}
  R_g(\widehat f_g)-R_g^{\star}
  =\operatorname{Est}_g(n)+\epsilon(g).
  \label{eq:supp-excess-decomposition}
\end{equation}

\paragraph{A1 (simultaneous certificates).}
Assume target-specific constants $A_g>0$, a shared function
$C_S:(0,1)\to(0,\infty)$, and exponents $\alpha(g)\in[1/2,1]$.
For every $n\geq1$ and
$\delta\in(0,1)$,
\begin{equation}
  \Pr\!\left(
    \forall g\in\mathcal G:\
    \operatorname{Est}_g(n)
    \leq A_g C_S(\delta)n^{-\alpha(g)}
  \right)
  \geq1-\delta.
  \label{eq:supp-A1}
\end{equation}
The event is joint across all displayed targets. If the trajectory is selected
from data, the event is understood uniformly over the admissible complete
trajectories; unrelated pointwise bounds are insufficient.

For each expert, set $C_i\coloneqq C(t_i)$ and define the same-anchor proxy
\begin{equation}
  D_i\coloneqq D(S_0,t_i)
  =C_i\bigl[1+\lambda d_{\to}(S_0,t_i)\bigr].
  \label{eq:supp-same-anchor-proxy}
\end{equation}
Define the aggregated-target proxy by
\begin{equation}
  D_{\mathrm{agg}}
  \coloneqq \max_{1\leq i\leq K}D_i+\Gamma(\mathcal T),
  \qquad \Gamma(\mathcal T)\geq0.
  \label{eq:supp-aggregate-proxy}
\end{equation}
This is a proxy definition rather than an empirical estimate or a universal
characterization of aggregation. The nonnegative term $\Gamma(\mathcal T)$
encodes the simultaneous-target burden of the selected certificate family.

\paragraph{A2 (proxy-to-exponent monotonicity).}
Set $D(g_i)=D_i$ and $D(g_{\mathrm{agg}})=D_{\mathrm{agg}}$. Within the
selected certificate family, assume
\begin{equation}
  \begin{aligned}
    D(g_a)\leq D(g_b)
    &\quad\Longrightarrow\quad
    \alpha(g_a)\geq\alpha(g_b),\\
    &\hspace{3.5em} g_a,g_b\in\mathcal G.
  \end{aligned}
  \label{eq:supp-A2}
\end{equation}
A2 is an explicit property of the certificate family, not a universal learning
law. Because $D_i\leq D_{\mathrm{agg}}$ for every $i$, it gives
\begin{equation}
  \begin{aligned}
    \alpha_{\mathrm{prog}}
    &\coloneqq \min_{1\leq k\leq K}\alpha(g_{\pi(k)})\\
    &=\min_{1\leq i\leq K}\alpha(g_i)
    \geq \alpha(g_{\mathrm{agg}})
    \coloneqq\alpha_{\mathrm{agg}}.
  \end{aligned}
  \label{eq:supp-exponent-order}
\end{equation}
This conclusion depends only on visiting every expert exactly once; it neither
distinguishes complete permutations nor proves that the greedy trajectory
minimizes a directed path objective.

\subsection{Finite-sample comparison and proof}

For a complete trajectory $\pi$, define
\begin{align}
  A_{\mathrm{prog}}
  &\coloneqq \max_{1\leq k\leq K}A_{g_{\pi(k)}},
  &
  \epsilon_{\mathrm{prog}}
  &\coloneqq \sum_{k=1}^{K}\epsilon(g_{\pi(k)}),
  \label{eq:supp-progressive-constants}\\
  A_{\mathrm{agg}}
  &\coloneqq A_{g_{\mathrm{agg}}},
  &
  \epsilon_{\mathrm{agg}}
  &\coloneqq \epsilon(g_{\mathrm{agg}}).
  \label{eq:supp-aggregate-constants}
\end{align}
The two proxy certificates are
\begin{align}
  B_{\mathrm{prog}}(n)
  &\coloneqq
  A_{\mathrm{prog}}K C_S(\delta)n^{-\alpha_{\mathrm{prog}}}
  +\epsilon_{\mathrm{prog}},
  \label{eq:supp-Bprog}\\
  B_{\mathrm{agg}}(n)
  &\coloneqq
  A_{\mathrm{agg}}C_S(\delta)n^{-\alpha_{\mathrm{agg}}}
  +\epsilon_{\mathrm{agg}}.
  \label{eq:supp-Bagg}
\end{align}

\paragraph{Theorem 1 (conditional proxy-certificate comparison, restated).}
Fix $n\geq1$ and $\delta\in(0,1)$. Under the ordered-head setup and A1--A2,
let
$\Delta\alpha\coloneqq
\alpha_{\mathrm{prog}}-\alpha_{\mathrm{agg}}\geq0$.
If
\begin{equation}
  A_{\mathrm{prog}}K
  \leq A_{\mathrm{agg}}n^{\Delta\alpha},
  \qquad
  \epsilon_{\mathrm{prog}}\leq\epsilon_{\mathrm{agg}},
  \label{eq:supp-finite-sample-conditions}
\end{equation}
then, on the joint event in A1,
\begin{equation}
  B_{\mathrm{prog}}(n)\leq B_{\mathrm{agg}}(n).
  \label{eq:supp-certificate-comparison}
\end{equation}

\paragraph{Proof.}
Equation \eqref{eq:supp-aggregate-proxy} and A2 imply
Eq.~\eqref{eq:supp-exponent-order}. On the event in A1, summing the $K$
target-specific progressive certificates and using
$\alpha(g_{\pi(k)})\geq\alpha_{\mathrm{prog}}$ with $n\geq1$ yields the
estimation component
$A_{\mathrm{prog}}K C_S(\delta)n^{-\alpha_{\mathrm{prog}}}$.
The aggregated estimation component is
$A_{\mathrm{agg}}C_S(\delta)n^{-\alpha_{\mathrm{agg}}}$.
Multiplying the first condition in
Eq.~\eqref{eq:supp-finite-sample-conditions} by the positive quantity
$C_S(\delta)n^{-\alpha_{\mathrm{prog}}}$ proves that the progressive
estimation component is no larger. Adding the second condition proves
Eq.~\eqref{eq:supp-certificate-comparison}. \hfill$\square$

If $\Delta\alpha>0$, the finite-sample constant condition is satisfied whenever
\begin{equation}
  n\geq n_0
  \coloneqq
  \max\!\left\{
    1,
    \left(\frac{A_{\mathrm{prog}}K}{A_{\mathrm{agg}}}\right)^{1/\Delta\alpha}
  \right\}.
  \label{eq:supp-sample-threshold}
\end{equation}
If $\Delta\alpha=0$, it instead reduces to
$A_{\mathrm{prog}}K\leq A_{\mathrm{agg}}$; increasing $n$ cannot overcome an
unfavorable constant ratio.

The theorem compares constructed proxy certificates for distinct
target-specific risks. The progressive quantity is a sum of $K$ target-wise
certificates and is not the excess risk of the final carrier under a single
common population risk. Accordingly, the result is not an unconditional
comparison of actual detection errors or of two algorithms evaluated under
one common risk. It is conditional on A1, A2, the finite-sample constant
condition, and the approximation-burden condition.

\section{Planning-Row Fidelity}
\label{sec:planning-row-fidelity}

This section gives the sufficient latent-cost fidelity-and-margin condition
used to justify the offline planning anchor. Let
\begin{equation}
  \begin{aligned}
    a_0
    &\coloneqq S_0,
    \qquad
    a_{k-1}\coloneqq t_{\pi(k-1)}\quad(k>1),\\
    \mathcal I_{k-1}
    &\coloneqq
    \{1,\ldots,K\}\setminus\\[-0.2em]
    &\qquad
    \{\pi(j):1\leq j<k\}.
  \end{aligned}
  \label{eq:supp-planning-sets}
\end{equation}
For every $i\in\mathcal I_{k-1}$, define the actual carrier-aware cost and the
offline planning cost by
\begin{equation}
  D_i^{(k)}\coloneqq D(S_{k-1},t_i),
  \qquad
  \widetilde D_i^{(k)}\coloneqq D(a_{k-1},t_i).
  \label{eq:supp-actual-offline-costs}
\end{equation}
Assume the planning-row fidelity condition
\begin{equation}
  \begin{aligned}
    \sup_{i\in\mathcal I_{k-1}}
    \left|
      d_{\to}(S_{k-1},t_i)-d_{\to}(a_{k-1},t_i)
    \right|
    &\leq\varepsilon_{k-1},\\
    \varepsilon_0&=0.
  \end{aligned}
  \label{eq:supp-planning-row-condition}
\end{equation}
This condition does not require $S_{k-1}=a_{k-1}$; it requires only similar
outgoing compatibility profiles over the remaining experts. By
Eq.~\eqref{eq:supp-latent-idtd}, it implies
\begin{equation}
  \left|D_i^{(k)}-\widetilde D_i^{(k)}\right|
  \leq \lambda C_i\varepsilon_{k-1},
  \qquad i\in\mathcal I_{k-1}.
  \label{eq:supp-planning-cost-perturbation}
\end{equation}

Let
\begin{equation}
  i^{\star}
  \in
  \operatorname*{arg\,min}_{i\in\mathcal I_{k-1}}
  \widetilde D_i^{(k)}
  \label{eq:supp-offline-minimizer}
\end{equation}
be an expert selected under the offline anchor.

\paragraph{Proposition (planning-row fidelity).}
Suppose that, for every
$j\in\mathcal I_{k-1}\setminus\{i^{\star}\}$,
\begin{equation}
  \widetilde D_j^{(k)}-\widetilde D_{i^{\star}}^{(k)}
  >
  \lambda\bigl(C_j+C_{i^{\star}}\bigr)\varepsilon_{k-1}.
  \label{eq:supp-planning-margin}
\end{equation}
Then $i^{\star}$ is the unique minimizer of the actual carrier-aware costs:
\begin{equation}
  \operatorname*{arg\,min}_{i\in\mathcal I_{k-1}}
  D(S_{k-1},t_i)
  =\{i^{\star}\}.
  \label{eq:supp-same-selection}
\end{equation}

\paragraph{Proof.}
For any $j\neq i^{\star}$, Eq.~\eqref{eq:supp-planning-cost-perturbation}
gives
\begin{align}
  D_j^{(k)}-D_{i^{\star}}^{(k)}
  &\geq
  \widetilde D_j^{(k)}-\widetilde D_{i^{\star}}^{(k)}
  -\lambda\bigl(C_j+C_{i^{\star}}\bigr)\varepsilon_{k-1}
  \notag\\
  &>0,
  \label{eq:supp-planning-proof}
\end{align}
where the strict inequality follows from
Eq.~\eqref{eq:supp-planning-margin}. Hence every remaining candidate has
strictly larger actual cost than $i^{\star}$. \hfill$\square$

At the first stage, $a_0=S_0$, so the offline and actual anchors coincide.
The proposition concerns the latent cost $D$. Connecting it to the implemented
table $\widehat D$ additionally requires anchor-wise row consistency. For
candidate pairs whose latent margin exceeds empirical estimation and
adaptor-optimization noise, a sufficient form is
\begin{equation}
  D(A,t_i)<D(A,t_j)
  \quad\Longrightarrow\quad
  \widehat D(A,t_i)<\widehat D(A,t_j).
  \label{eq:supp-empirical-row-consistency}
\end{equation}
The empirical score is not required to satisfy the directed triangle
inequality.


\end{document}